\documentclass[]{fairmeta}

\usepackage{caption}
\usepackage{amssymb, bbm}
\usepackage{inconsolata}
\usepackage{color,colortbl}
\newcolumntype{C}{D>{\;/\;}{3.3}}

\usepackage{dsfont}  %

\definecolor{color1}{HTML}{006EB8}
\definecolor{color2}{HTML}{009B55}
\definecolor{color3}{HTML}{00A99A}
\definecolor{color4}{HTML}{3C8031}
\definecolor{color5}{HTML}{006795}
\definecolor{color6}{HTML}{00AEB3}
\definecolor{mygray}{gray}{0.9}
\definecolor{azureblue}{rgb}{0,0.5,1}
\makeatletter
\g@addto@macro\bfseries{\boldmath}
\makeatother

\usepackage{hyperref}
\hypersetup{
  colorlinks   = true, %
  urlcolor     = magenta, %
  linkcolor    = magenta, %
  citecolor   = azureblue %
}
\usepackage{url}
\usepackage{booktabs}
\usepackage{algpseudocodex}
\definecolor{comm}{gray}{0.5}
\usepackage{algorithm}
\usepackage{multirow}
\usepackage{enumitem}
\usepackage{xcolor,colortbl}
\usepackage{wrapfig,lipsum,booktabs}
\usepackage{bbm}
\usepackage{tikz}
\usepackage{ifthen}
\usepackage{mathtools}
\usepackage{listings,multicol}
\definecolor{light-gray}{gray}{0.9}

\usepackage{amsmath}
\usepackage{amsfonts}
\usepackage{amssymb}
\usepackage{calligra}
\usepackage{calrsfs}

\newcommand{\by}{\mathbf{y}}
\newcommand{\bx}{\mathbf{x}}

\definecolor{blanchedalmond}{rgb}{1.0, 0.92, 0.8}
\definecolor{carmine}{rgb}{0.59, 0.0, 0.09}
\definecolor{lightblue}{rgb}{0.22,0.45,0.70}%

\renewcommand{\mathbf}{\boldsymbol}

\makeatletter
\def\Ddots{\mathinner{\mkern1mu\raise\p@
\vbox{\kern7\p@\hbox{.}}\mkern2mu
\raise4\p@\hbox{.}\mkern2mu\raise7\p@\hbox{.}\mkern1mu}}
\makeatother

\newcommand{\Rex}{\mathrm{Regex}}
\newcommand{\Prm}{\mathrm{Prm}}

\makeatletter
\newcommand{\subalign}[1]{%
  \vcenter{%
    \Let@ \restore@math@cr \default@tag
    \baselineskip\fontdimen10 \scriptfont\tw@
    \advance\baselineskip\fontdimen12 \scriptfont\tw@
    \lineskip\thr@@\fontdimen8 \scriptfont\thr@@
    \lineskiplimit\lineskip
    \ialign{\hfil$\m@th\scriptstyle##$&$\m@th\scriptstyle{}##$\hfil\crcr
      #1\crcr
    }%
  }%
}
\makeatother

\usepackage{arydshln}

\usepackage[most]{tcolorbox}
\usepackage{colortbl}
\usepackage{nicematrix}  %
\crefname{figure}{Figure}{Figures}

\title{\method{}: Optimizing Mathematical Reasoning through Stepwise Binary Feedback}

\author[1,2]{Yen-Ting Lin}
\author[1]{Di Jin}
\author[1]{Tengyu Xu}
\author[3,4]{Tianhao Wu}
\author[3]{Sainbayar Sukhbaatar}
\author[1]{Chen Zhu}
\author[1]{Yun He}
\author[2]{\\Yun-Nung Chen}
\author[3]{Jason Weston}
\author[3]{Yuandong Tian}
\author[1]{Arash Rahnama}
\author[1]{Sinong Wang}
\author[1]{Hao Ma}
\author[1]{Han Fang}

\affiliation[1]{Meta GenAI}
\affiliation[2]{National Taiwan University}
\affiliation[3]{Meta FAIR}
\affiliation[4]{UC Berkeley}

\newcommand{\method}[0]{\textsc{Step-KTO}}

\abstract{
Large language models (LLMs) have recently demonstrated remarkable success in mathematical reasoning. Despite progress in methods like chain-of-thought prompting and self-consistency sampling, these advances often focus on final correctness without ensuring that the underlying reasoning process is coherent and reliable. This paper introduces \method{}, a training framework that combines process-level and outcome-level binary feedback to guide LLMs toward more trustworthy reasoning trajectories. By providing binary evaluations for both the intermediate reasoning steps and the final answer, \method{} encourages the model to adhere to logical progressions rather than relying on superficial shortcuts. Our experiments on challenging mathematical benchmarks show that \method{} significantly improves both final answer accuracy and the quality of intermediate reasoning steps. For example, on the MATH-500 dataset, \method{} achieves a notable improvement in Pass@1 accuracy over strong baselines. These results highlight the promise of integrating stepwise process feedback into LLM training, paving the way toward more interpretable and dependable reasoning capabilities.
}

\date{\today}
\correspondence{Yen-Ting Lin (\email{ytl@ieee.org}), Di Jin(\email{jindi@meta.com})}

\begin{document}

\maketitle

\section{Introduction}
\label{section:intro}

\begin{figure}[t]
    \centering
    \includegraphics[trim={11cm 13.7cm 9.5cm 12cm}, clip,width=\linewidth]{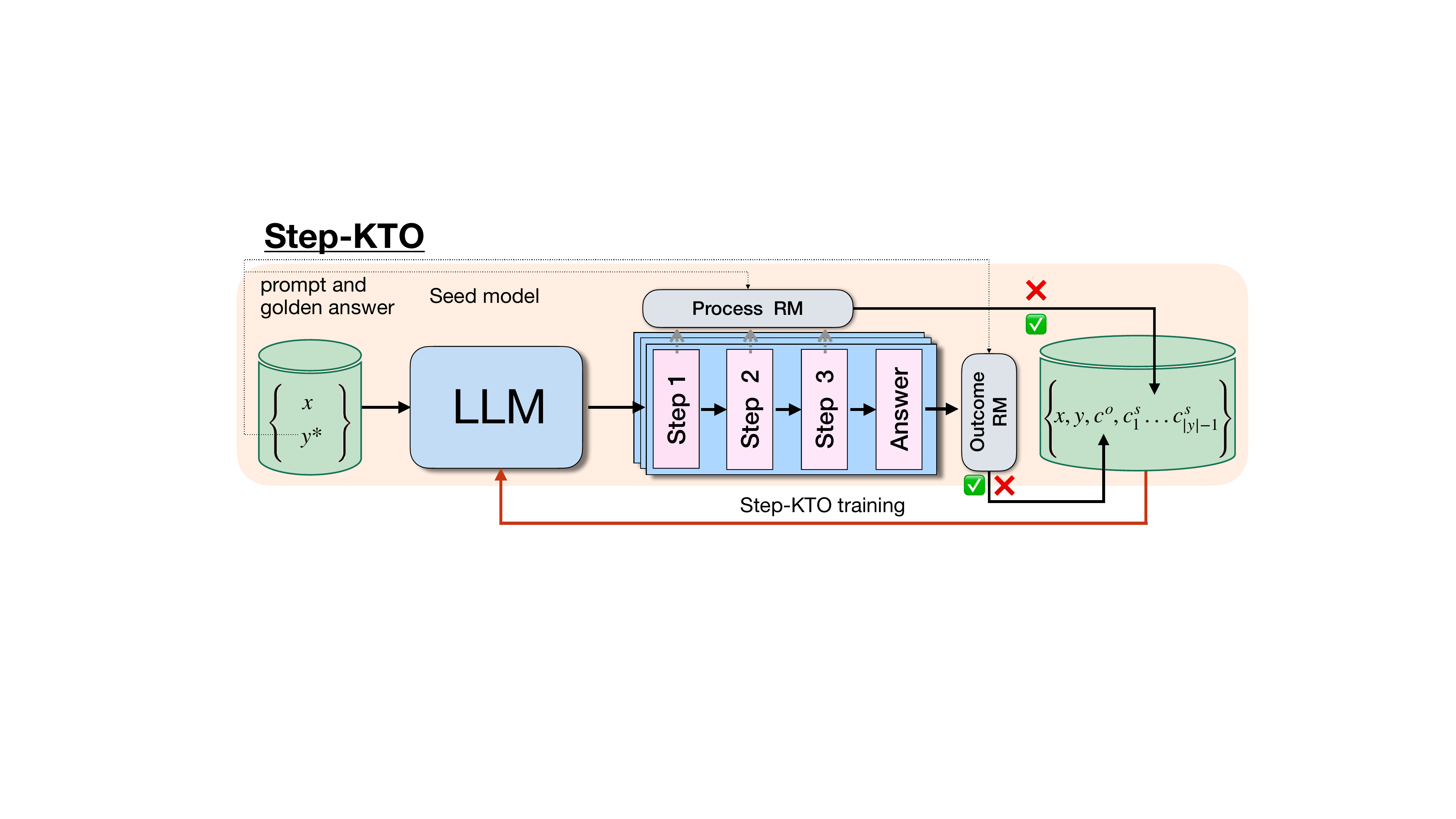}
    \caption{\textbf{\method{} Training Process.} 
    Given a dataset of math problems (left), a language model (LLM) produces both reasoning steps and a final answer. 
    Each intermediate reasoning step is evaluated by a process reward model (Process RM), and the final answer is assessed by an outcome reward model (Outcome RM).
    The binary feedback signals from both levels (outcome-level correctness $c^o$ and stepwise correctness $c^s_h$) are recorded together with the input $(x)$ and the model's response $(y)$ \S\ref{subsec:problem_setup}.
    These signals are then used to compute the \method{} loss, guiding the LLM to not only produce correct final answers but also maintain coherent and correct reasoning steps \S\ref{subsec:incorporating_stepwise_feedback}.
    Through multiple iterations of this training process \S\ref{subsec:training_process}, the model progressively improves both its stepwise reasoning and final answer accuracy.
    }
    \label{fig:method}
    \vspace{-0.5cm}
\end{figure}

Large language models (LLMs) have recently shown remarkable capabilities in reasoning-intensive tasks such as coding \citep{humaneval_code_markchen, doi:10.1126/science.abq1158, codellama} and solving complex mathematical problems \citep{DBLP:journals/corr/abs-2402-03300, DBLP:conf/iclr/AzerbayevSPSMJD24}. Prompting strategies like chain-of-thought prompting \citep{cot_first_work, wei2022chain, kojima2022large, DBLP:conf/emnlp/Adolphs0USW22} and self-consistency sampling \citep{wang2022selfconsistency} enhance these models’ final-answer accuracy by encouraging them to articulate intermediate reasoning steps. However, a significant issue remains: even when these methods boost final-answer correctness, the internal reasoning steps are often unreliable or logically inconsistent \citep{orm_prm_training, verify_step_by_step_prm_first_work}.

This discrepancy between correct final answers and flawed intermediate reasoning limits our ability to trust LLMs in scenarios where transparency and correctness of each reasoning stage are crucial \citep{measuring_faithfulness}. For example, in mathematical problem-solving, a model might produce the right answer for the wrong reasons \citep{faithful_cot, zheng2024processbench}, confounding our understanding of its true capabilities \citep{unfaithful_explanation_cot}. To address this, researchers are increasingly emphasizing the importance of guiding models to produce not just correct final answers, but also verifiable and faithful step-by-step solution paths \citep{orm_prm_training,DBLP:journals/corr/abs-2402-03300,rewarding_progress}.

Prior work in finetuning has largely focused on outcome-level correctness, using outcome reward models to improve the probability of final-answer accuracy \citep{orm_first, v_star, gen_orm}. While effective, such an approach does not ensure that the intermediate reasoning steps are valid. 
Conversely, while process-level supervision through process reward models (PRMs) \citep{verify_step_by_step_prm_first_work, math_shepherd, omega_prm} can guide models to follow correct reasoning trajectories, prior work has mainly used PRMs as a ranking method rather than a way to provide stepwise feedback. As a result, relying solely on process-level supervision may lead models to prioritize step-by-step correctness without guaranteeing a correct final outcome.

In this paper, we introduce Stepwise Kahneman-Tversky-inspired Optimization (\method{}), a training framework that integrates both process-level and outcome-level binary feedback to produce coherent and correct reasoning steps alongside high-quality final answers. Our approach evaluates each intermediate reasoning step against known correct patterns using a PRM, while simultaneously leveraging a rule-based reward signal for the final answer. To fuse these signals, we employ a Kahneman-Tversky-inspired value function \citep{kt, kto} that emphasizes human-like risk and loss aversion, encouraging models to gradually correct their reasoning and avoid errors. The result is a training objective that aligns the entire reasoning trajectory with verified solutions while ensuring that final correctness remains a top priority.

Figure~\ref{fig:method} illustrates the \method{} pipeline. We start with a base LLM and repeatedly refine it through iterative training. At each iteration, the PRM provides step-level binary feedback that helps the model navigate correct solution paths, while the outcome-level binary feedback ensures that the final answer is correct. The Kahneman-Tversky-inspired value function transforms these binary signals into guidance that progressively reduces errors in the chain-of-thought. Over successive rounds, \method{} yields systematically more accurate intermediate reasoning steps and steadily improves the final-answer accuracy.

We evaluate \method{} on challenging mathematical reasoning benchmarks including MATH-500 \citep{MATH_OG, verify_step_by_step_prm_first_work}, AMC23 \citep{AMC2023}, and AIME24\citep{AIME2024}. Our experiments show that incorporating both process-level and outcome-level signals leads to substantial improvements over state-of-the-art baselines that rely solely on final-answer supervision.
For example, on MATH-500, \method{} improves Pass@1 accuracy from 53.4\% to 63.2\%, while also producing more coherent and trustworthy step-by-step reasoning. Moreover, iterative training with \method{} achieves cumulative gains, demonstrating that balancing process- and outcome-level feedback refines reasoning quality over time.

In summary, our key contributions are:
\begin{itemize}[leftmargin=5mm]
	\item We propose \method{}, a novel finetuning framework that combines process-level and outcome-level feedback, encouraging both correct final answers and faithful step-by-step reasoning.
    \item We show that iterative training with \method{} yields consistent cumulative improvements, showing the effectiveness of combined process-level and outcome-level feedback in refining LLM reasoning.
	\item We demonstrate that \method{} surpasses state-of-the-art baselines on multiple math reasoning tasks, delivering higher accuracy (63.2\% vs 53.4\% Pass@1 on MATH-500) and more reliable intermediate solutions.
\end{itemize}

\section{Methodology}
\label{sec:methodology}

\subsection{Problem Setup and Notation}
\label{subsec:problem_setup}

We adopt the notation and setup similar to \citet{rewarding_progress}. Let $\mathcal{D}=\{(\bx_i, \by_{\bx_i}^{\star})\}_{i}$ be a dataset of math problems, where each problem $\bx \in \mathcal{X}$ has an associated ground-truth solution sequence $\by_{\bx}^{\star} = (s_1^{\star}, s_2^{\star}, \ldots, s_{|\by^{\star}|}^{\star}) \in \mathcal{Y}$. A policy model $\pi_{\theta}$, parameterized by $\theta$, generates a response sequence $\by = (s_1, s_2, \ldots, s_{|\by|})$ autoregressively given the problem $\bx$, where each step $s_h$ is a reasoning step separated by a special token (e.g., "\#\# Step").

The correctness of the final answer $\by$ can be automatically determined by a rule-based correctness function \(\Rex(\by,\by_{\bx}^{\star})\in \{0,1\}\), which compares the model’s final derived answer to the ground-truth final answer \citep{MATH_OG}. The model’s final answer is explicitly denoted using a special format in the final step $s_{|\by|}$, such as \texttt{\text{\\boxed\{$\cdot$\}}}, allowing the correctness function to easily extract and verify it. Our primary objective is to improve the expected correctness of the final answer:
\[
\mathbb{E}_{\bx \in \mathcal{D}, \ \by \sim \pi_{\theta}(\cdot \mid \bx)}[\Rex(\by,\by_{\bx}^{\star})].
\]

Ensuring a correct final answer does not guarantee logically sound intermediate reasoning. To address this, we incorporate a stepwise binary correctness signal \(\Prm(\bx, \by_{\bx}^{\star}, s_h)\in\{0,1\}\) for each reasoning step \(s_h\). Unlike the final-answer correctness \(\Rex\), this signal directly measures whether each intermediate step is locally valid and aligns with proper problem-solving principles, without strictly mirroring the reference solution steps. We obtain these stepwise correctness evaluations by prompting an LLM (\texttt{Llama-3.1-70B-Instruct}) as our process reward model (PRM), following the structured template in Appendix \ref{app:stepwise_feedback_prompt}.

In summary, we have two levels of binary signals:
\begin{itemize}[leftmargin=5mm]
    \item \textbf{Outcome feedback}: $\Rex(\by, \by_{\bx}^{\star}) \in \{0,1\}$ indicates if the final answer derived from $\by$ is correct.
    \item \textbf{Stepwise feedback}: $\Prm(\bx, \by_{\bx}^{\star}, s_h) \in \{0,1\}$ indicates if the intermediate reasoning step $s_h$ is correct.
\end{itemize}

Our goal is to integrate both of these signals into the training objective of $\pi_{\theta}$. By doing so, we guide the model to produce not only correct final answers but also to maintain correctness, coherence, and reliability throughout its reasoning trajectory. This integrated approach will be formalized through the \method{} framework.

\subsection{KTO Background}
\label{subsec:kto_background}

KTO \citep{kto} aims to align a policy $\pi_\theta$ with binary feedback using a Kahneman-Tversky-inspired value function \citep{kt}. Rather than maximizing the log-likelihood of preferred outputs or directly using reinforcement learning, KTO defines a logistic value function that is risk-averse for gains and risk-seeking for losses.

The original KTO loss focuses on the final-answer level. Let:
\begin{align*}
r_\theta(x,y) &= \log \frac{\pi_\theta(y\mid x)}{\pi_{\text{ref}}(y\mid x)},\\
z_0 &= \text{KL}\bigl(\pi_\theta(y'\mid x)\parallel \pi_{\text{ref}}(y'\mid x)\bigr),\\
v(x,y) &= \begin{cases}
\lambda_D \,\sigma\bigl(\beta (r_\theta(x,y) - z_0)\bigr) & \text{if } \Rex(\by, \by_{\bx}^{\star})=1, \\[6pt]
\lambda_U \,\sigma\bigl(\beta (z_0 - r_\theta(x,y))\bigr) & \text{if } \Rex(\by, \by_{\bx}^{\star})=0.
\end{cases}
\end{align*}
Here, $\pi_{\text{ref}}$ is a reference policy (typically the initial model checkpoint) that provides a baseline for comparison, $\sigma$ is the logistic function, $\beta > 0$ controls risk aversion, and $\lambda_D,\lambda_U$ are weighting coefficients. The $z_0$ term, where $y'$ denotes an arbitrary output sequence, serves as a reference point to ensure balanced optimization. The KTO loss at the outcome level is:
\begin{equation}
L_{\text{KTO}}(\pi_\theta, \pi_{\text{ref}}) \;=\; \mathbb{E}_{x,y \sim D}[\lambda_y - v(x,y)],
\label{eq:kto_loss}
\end{equation}
where $\lambda_y = \lambda_D$ if $\Rex(\by, \by_{\bx}^{\star})=1$ and $\lambda_y = \lambda_U$ if $\Rex(\by, \by_{\bx}^{\star})=0$.

\subsection{\method{}}
\label{subsec:incorporating_stepwise_feedback}
While KTO ensures correctness of final answers, many reasoning tasks require validity at each intermediate step. We extend KTO by incorporating stepwise binary feedback $\Prm(\bx, \by_{\bx}^{\star}, s_h)$ to assess the quality of each reasoning step.
We begin by defining an \emph{implied reward} at the step level:
\[
r_\theta(x,s_h) = \log \frac{\pi_\theta(s_h \mid x,s_{<h})}{\pi_{\text{ref}}(s_h \mid x,s_{<h})}.
\]
This quantity can be viewed as the incremental advantage of producing step $s_h$ under $\pi_\theta$ compared to $\pi_{\text{ref}}$. It captures how much more (or less) reward is implied by choosing $s_h$ over the reference model’s baseline likelihood, conditioned on the same context $(x,s_{<h})$.
Next, we introduce a stepwise KL baseline:
\[
z_0^{(step)} = \text{KL}\bigl(\pi_\theta(s_h' \mid x,s_{<h}') \parallel \pi_{\text{ref}}(s_h' \mid x,s_{<h}')\bigr).
\]
Analogous to $z_0$ at the outcome level, $z_0^{(step)}$ serves as a local reference point. It prevents the model from gaining reward merely by diverging from the reference and ensures that improvements are grounded in genuine reasoning quality.
Given the binary stepwise feedback $\Prm(\bx, \by_{\bx}^{\star}, s_h)$, we define a value function that parallels the outcome-level case.
If a step $s_h$ is deemed stepwise-desirable, the model should increase its implied reward $r_\theta(x,s_h)$ relative to $z_0^{(step)}$~\citep{huang-chen-2024-factalign}. 
Conversely, if $s_h$ is stepwise-undesirable, the model is encouraged to lower that implied reward. Formally:
\[
v^{(step)}(x,s_h) = \begin{cases}
\lambda_D^{(step)}\,\sigma\bigl(\beta_{step}(r_\theta(x,s_h)-z_0^{(step)})\bigr) & \text{if } \Prm(\bx, \by_{\bx}^{\star}, s_h)=1, \\[6pt]
\lambda_U^{(step)}\,\sigma\bigl(\beta_{step}(z_0^{(step)}-r_\theta(x,s_h))\bigr) & \text{if } \Prm(\bx, \by_{\bx}^{\star}, s_h)=0.
\end{cases}
\]
Here, $\lambda_D^{(step)},\lambda_U^{(step)}$ and $\beta_{step}$ mirror their outcome-level counterparts, controlling the strength of the reward or penalty at the granularity of individual steps. By leveraging these signals, the stepwise value function $v^{(step)}$ directs the model’s distribution toward steps deemed correct and coherent, and away from those that are not.
With these definitions, the stepwise loss is:
\[
\mathcal{L}_\text{step}(\pi_\theta,\pi_{\text{ref}}) = \mathbb{E}_{x,y,s_h \sim D^{(step)}}[\lambda_y^{(step)} - v^{(step)}(x,s_h)].
\]
where $\lambda_y^{(step)} = \lambda_D^{(step)}$ if $\Prm(\bx, \by_{\bx}^{\star}, s_h)=1$ and $\lambda_y^{(step)} = \lambda_U^{(step)}$ if $\Prm(\bx, \by_{\bx}^{\star}, s_h)=0$.

Combining the stepwise objective with the outcome-level KTO loss (Eq.~\ref{eq:kto_loss}) yields the final \method{} objective:
\begin{equation}
\mathcal{L}_{\text{\method{}}}(\pi_\theta,\pi_{\text{ref}}) = \mathcal{L}_{\text{KTO}}(\pi_\theta,\pi_{\text{ref}}) + \mathcal{L}_\text{step}(\pi_\theta,\pi_{\text{ref}}). 
\label{eq:step_kto_loss}
\end{equation}
This composite loss encourages the model to produce not only correct final answers but also to refine each intermediate step. By jointly optimizing outcome-level and stepwise-level feedback, \method{} ensures that the model’s entire reasoning trajectory—from the earliest steps to the final solution—is both correct and coherent.

\subsection{Iterative Training}
\label{subsec:training_process}

We train our models using an iterative procedure inspired by previous alignment methods that refine a model's parameters over multiple rounds \citep{star, self_rewarding_lm, irpo, self_consistency_preference_optimization}.
For \texttt{Llama-3.3-70B-Instruct}, we use it directly as our seed model $M_0$. For Llama-3.1 models, we first perform supervised finetuning on the training data before using them as $M_0$. Starting from $M_0$, we refine it iteratively to obtain $M_1, M_2, \ldots, M_T$ using the following procedure:

\begin{enumerate}[leftmargin=10pt]
    \item \textbf{Candidate Generation:} For each problem $\bx \in \mathcal{D}$, we sample 8 candidate solutions $\by^k \sim \pi_{M_t}(\cdot \mid \bx)$ using temperature $T=0.7$ and nucleus sampling with $p=0.95$ \citep{nucleus_sampling}. This stochastic decoding strategy encourages diverse candidate solutions, aiding both positive and negative sample selection.

    \item \textbf{Outcome Assessment:} We evaluate each candidate $\by^k$ against the ground-truth solution $\by_{\bx}^{\star}$ using the outcome correctness function $\Rex(\by^k, \by_{\bx}^{\star})$. 
    If no sampled solutions are correct, we include the ground-truth solution $\by_{\bx}^{\star}$ as a positive sample, as suggested by \citet{irpo}.
    If all sampled solutions are correct, we discard this problem in the current iteration to prioritize learning from problems where the model can still improve.
    
    \item \textbf{Stepwise Evaluation:} For the selected solutions, we apply the stepwise correctness function $\Prm(\bx, \by_{\bx}^{\star}, s_h)$ to assess the quality of each reasoning step. This yields a set of binary signals indicating whether each intermediate step aligns with desirable reasoning patterns.

    \item \textbf{Dataset Construction:} We aggregate these annotated samples into
    $
    \mathcal{D}_{M_t} = \{(\bx, \by, c^{out}, c^{step}_1, \ldots, c^{step}_{S-1}) \mid \by \in \mathcal{D}\},
    $
    where $c^{out} = \Rex(\by,\by_{\bx}^{\star})$ is the outcome-level correctness, and $c^{step}_h = \Prm(\bx, \by_{\bx}^{\star}, s_h)$ are the stepwise correctness indicators for the $S-1$ intermediate steps of the solution $\by$.
    \footnote{At each iteration $t$, the dataset $\mathcal{D}_{M_t}$ is constructed specifically from $M_t$. Thus, $M_1$ is trained on the dataset derived from seed model $M_0$ shared by all methods, $M_2$ on the dataset derived from $M_1$ specifically for method testing, and so forth.}

    \item \textbf{Parameter Update:} Using $\mathcal{D}_{M_t}$, we update the model parameters according to the chosen alignment objective—either our \method{} loss or a baseline method (e.g., IRPO).

    \item \textbf{Iteration:} We repeat this process for $T$ iterations, each time producing a new model $M_{t+1}$ refined from $M_t$.
\end{enumerate}

While KTO and \method{} does not inherently require balanced positive and negative samples, we impose this constraint for fairness when comparing against pairwise preference-based baselines like DPO. Specifically, we randomly sample at most two pairs per problem per iteration, ensuring a consistent number of training examples across different alignment strategies. This controlled sampling regime facilitates direct comparisons between methods and clarifies the impact of stepwise and outcome-level feedback on the model’s refinement process.

\section{Experiments}
\label{sec:results}

\subsection{Task and Datasets}

We evaluate our approach on established math reasoning benchmarks derived from high school competition-level exams. These tasks test the model's ability to solve challenging mathematical problems spanning various domains and difficulty levels. All problems require the model to produce a final answer, which is often a number, a simplified expression (e.g., $\frac{\pi}{2}$ or $1 \pm \sqrt{19}$), or a short textual response (e.g., ``east'').

\begin{itemize}[leftmargin=5mm]
    \item \textbf{MATH-500:} A curated subset of 500 problems drawn from the MATH dataset \citep{MATH_OG}, selected as in \citet{verify_step_by_step_prm_first_work}. These problems cover seven subjects: Prealgebra, Algebra, Number Theory, Counting and Probability, Geometry, Intermediate Algebra, and Precalculus, ensuring a broad evaluation of mathematical reasoning skills.
    \item \textbf{AMC23:} A test set of 40 problems from the American Mathematics Competitions (AMC 12, 2023)\footnote{\url{https://github.com/QwenLM/Qwen2.5-Math/blob/main/evaluation/data/amc23/test.jsonl}}. These problems are known for their subtlety and problem-solving depth, providing a stringent test of reasoning ability and accuracy.
    \item \textbf{AIME24:} A test set of 30 problems from the American Invitational Mathematics Examination (AIME, 2024)\footnote{\url{https://github.com/QwenLM/Qwen2.5-Math/blob/main/evaluation/data/aime24/test.jsonl}}. Each problem typically requires multiple steps of intricate reasoning, posing a higher-level challenge that further differentiates models based on their capacity to follow extended solution paths 
\end{itemize}

To evaluate the correctness of the model's outputs, we follow standard practices in mathematical LLM evaluation \citep{MATH_OG}. First, we parse the model’s generated solution using regular expressions to extract the final answer. Then, we employ \textsc{sympy}\footnote{\url{https://github.com/sympy/sympy}} to check for mathematical equivalence between the generated answer and the ground-truth solution. This approach ensures a fair comparison that accounts for minor stylistic or representational differences in the final answer.

We report results using two standard metrics:
\begin{itemize}[leftmargin=5mm]
    \item \textbf{Pass@1:} The ratio that a single greedy completion $\by$ from $\pi_{\theta}$ is correct.
    \item \textbf{Maj@8:} The accuracy obtained by generating 8 candidate solutions $\by^k \sim \pi_{\theta}(\cdot \mid \bx)$ at temperature $T=0.7$~\citep{temperature_sampling_old, temperature_sampling} and selecting the majority answer, as introduced by \citet{wang2022selfconsistency}\footnote{Pilot experiments indicated that varying the temperature within a reasonable range ($T=0.5-1.0$) had limited impact on overall Maj@8 performance.}.
\end{itemize}

These metrics reflect both direct accuracy under deterministic decoding (Pass@1) and the model’s robustness under multiple sampled solutions (Maj@8), providing a comprehensive assessment of model performance on challenging mathematical reasoning tasks.

In addition to these evaluation benchmarks, all experiments are conducted using a large-scale prompt set, $\mathcal{D}_{\text{Numina}}$, referred to as \texttt{NuminaMath} \citep{numina_math_datasets}. \texttt{NuminaMath} comprises a broad range of math problems and their solutions, spanning difficulty levels from elementary to high school competition standards. To ensure the integrity of final answers, we remove subsets of synthetic questions and Orca Math problems \citep{orca_math}, as their correctness are not verified by human.

\subsection{Baseline Methods}

We evaluate our proposed \method{} against several strong baseline approaches for mathematical reasoning. All methods are trained using offline iterative optimization, with online preference learning left as future work:

\begin{itemize}[leftmargin=5mm]
    \item \textbf{RFT (Rejection Finetuning)}: Following \citet{rejection_fine_tuning}, this method performs supervised finetuning on the filtered dataset $\{(\bx,\by) \in \mathcal{D}_{M_t} \mid c^{out}=1\}$, retaining only solutions with correct final answers. Unlike \method{}, RFT does not incorporate any explicit preference or reward signals, instead directly mimicking the ground-truth solutions.

    \item \textbf{IRPO (Iterative Reasoning Preference Optimization)}: Extending the ideas of DPO \citep{dpo}, IRPO applies iterative training with pairwise preferences at the outcome level, enhanced by an additional NLL loss term to stabilize training \citep{irpo}. IRPO does not utilize stepwise feedback, focusing solely on outcome correctness for model refinement.

    \item \textbf{KTO (Kahneman-Tversky Optimization):} KTO \citep{kto} applies a HALO-based loss derived from the Kahneman-Tversky value function (see \S\ref{subsec:kto_background}), instilling human-like risk aversion and asymmetric weighting of gains and losses. Like the other outcome-level methods, it does not incorporate stepwise correctness signals.

    \item \textbf{SimPO and IPO}: SimPO \citep{simpo} and IPO \citep{ipo} are both variants of DPO that optimize from pairwise preferences at the outcome level only. Unlike DPO, which relies on the Bradley-Terry model \citep{bradley1952rank} and can overemphasize deterministic preferences, SimPO and IPO apply simpler transformations to directly utilize preference probabilities. They primarily target safer, more stable optimization rather than introducing explicit mechanisms to enhance the reasoning performance.

    \item \textbf{Step-DPO}: A variant of DPO that optimizes stepwise preferences rather than outcome-level preferences \citep{step_dpo}. By identifying and correcting specific erroneous reasoning steps, Step-DPO aims to provide more granular supervision for long-chain reasoning tasks. However, it requires additional data processing to construct stepwise preference pairs and relies on rejection sampling to filter out incorrect intermediate steps.

\end{itemize}

\subsection{Implementation Details}
We use AdamW ($\beta_1=0.9$, $\beta_2=0.95$, weight decay $=0.1$) with a linear warmup for the first 100 steps and a cosine decay schedule that reduces the learning rate to $0.1 \times$ its initial value. The starting learning rate is $1.0 \times 10^{-6}$, and we apply global norm gradient clipping of 1.0. 
The effective global batch size is set to approximately one million tokens, and we train for about 2000 steps, periodically evaluating our models during training on the hold-out test set from MATH \citep{MATH_OG}\footnote{MATH-500 questions are excluded.} to select the best checkpoint for each method.
For IRPO, we use an NLL weight of 0.2. We set $\beta=0.1$ for all methods. All training jobs are run on 64 H100 GPUs. 

\subsection{Main Results}

\begin{table}[t]
    \small
    \centering
    \setlength{\tabcolsep}{7.0pt}
    \begin{NiceTabular}{lcccccc}
         \toprule
         {\bf Method} & \multicolumn{2}{c}{\bf MATH-500} & \multicolumn{2}{c}{\bf AMC23} & \multicolumn{2}{c}{\bf AIME24} \\
         & Pass@1 & Maj@8 & Pass@1 & Maj@8 & Pass@1 & Maj@8 \\
         \midrule
         \rowcolor{blue!7}
         \rowcolor{blue!7} {\small \hspace{-2mm} \emph{Llama-3.1-8B-Instruct} } &  & & &  &  & \\
         \rowcolor{blue!7} ~~Seed model $M_0$ & 53.4 & 55.0 & 35.0 & 37.5 & ~~3.3 & ~~6.7 \\ %
         \rowcolor{blue!7} ~~Rejection Finetuning $M_3$ & 53.8 & 56.0 & 30.0 & 32.5 & 10.0 & ~~6.7 \\ %
         \rowcolor{blue!7} ~~IRPO $M_3$ & 55.4 & 59.2 & 35.0 & 40.0 & ~~6.7 & ~~6.7 \\ %
         \rowcolor{blue!7} ~~KTO $M_3$ & 60.6 & 61.6 & 35.0 & 32.5 & \textbf{16.7} & \textbf{16.7} \\ %
         \rowcolor{blue!7} ~~\method{} (ours) $M_3$ & \textbf{63.2} & \textbf{64.6} & \textbf{47.5} & \textbf{47.5} & \textbf{16.7} & \textbf{16.7} \\ %
         \midrule
         \rowcolor{green!7} {\small \hspace{-2mm} \emph{Llama-3.1-70B-Instruct} } &  & & &  &  & \\
         \rowcolor{green!7} ~~Seed model $M_0$ & 74.6 & 76.2 & 40.0 & 60.0 & 13.3 & 16.7 \\
         \rowcolor{green!7} ~~Rejection Finetuning $M_1$ & 74.8 & 73.6 & 55.0 & 60.0 & 13.3 & 13.3 \\ %
         \rowcolor{green!7} ~~IRPO $M_1$ & 74.4 & 74.8 & 55.0 & 57.5 & 10.0 & 13.3 \\ %
         \rowcolor{green!7} ~~KTO $M_1$ & 75.6 & 77.2 & 55.0 & 65.0 & 13.3 & 13.3 \\ %
         \rowcolor{green!7} ~~\method{} (ours) $M_1$ & \textbf{76.2} & \textbf{78.4} & \textbf{60.0} & \textbf{67.5} & \textbf{16.7} & \textbf{20.0} \\ %
         \midrule
         \rowcolor{green!7} {\small \hspace{-2mm} \emph{Llama-3.3-70B-Instruct}} $M_0$ & 75.8 & 77.6 & 57.5 & 60.0 & 26.7 & 30.0 \\ %
         \rowcolor{green!7} ~~Rejection Finetuning $M_1$ & 77.4 & 78.4 & 60.0 & 65.0 & 20.0 & 23.3 \\ %
         \rowcolor{green!7} ~~IRPO $M_1$ & 78.6 & 80.8 & 55.0 & 57.5 & 23.3 & 26.7 \\ %
         \rowcolor{green!7} ~~KTO $M_1$ & 78.6 & 79.8 & 60.0 & 65.0 & 20.0 & 23.3 \\ %
         \rowcolor{green!7} ~~\method{} (ours) $M_1$ & \textbf{79.6} & \textbf{81.6} & \textbf{70.0} & \textbf{75.0} & \textbf{30.0} & \textbf{33.3} \\
         \midrule
         \rowcolor{gray!7} \hspace{-2mm} Llama-3.1-8B-Instruct & 51.4 & 55.2 & 15.0 & 27.5 & 3.3 & 3.3 \\ %
         \rowcolor{gray!7} \hspace{-2mm} Llama-3.1-70B-Instruct & 64.8 & 70.4 & 37.5 & 47.5 & 10.0 & 30.0 \\ %
         \rowcolor{gray!7} \hspace{-2mm} Llama-3.1-405B-Instruct & 68.8 & 74.4 & 47.5 & 52.5 & 30.0 & 26.6 \\
         \rowcolor{gray!7} \hspace{-2mm} O1 & 94.8 & - & - & - & 78.0 & - \\
         \rowcolor{gray!7} \hspace{-2mm} O1-Mini & ~90.0 & - & 90.0 & 90.0 & 33.3 & 46.7 \\
         \rowcolor{gray!7} \hspace{-2mm} Gemini 1.5 Pro& 79.4 & 83.0 & 75.0 & 82.5 & 26.7 & 26.7 \\
         \rowcolor{gray!7} \hspace{-2mm} GPT-4o & 73.0 & 76.4 & 57.5 & 70.0 & 10.0 & 16.7 \\
         \rowcolor{gray!7} \hspace{-2mm} Claude 3.5 Sonnet & 70.0 & 74.4 & 62.5 & 67.5 & 23.3 & 26.7 \\
         \rowcolor{gray!7} \hspace{-2mm} Grok-Beta & 67.0 & 72.2 & 50.0 & 52.5 & 10.0 & 13.3 \\
         \bottomrule
    \end{NiceTabular}
    \caption{{\bf Math problem solving performance} comparing Llama models of different sizes and proprietary models. Results show accuracy on MATH-500, AMC23, and AIME24 test sets using both greedy decoding (Pass@1) and majority voting over 8 samples (Maj@8). Models highlighted in \setlength{\fboxsep}{2pt}\colorbox{blue!7}{blue} are 8B parameter models, \setlength{\fboxsep}{2pt}\colorbox{green!7}{green} are 70B parameter models, and \setlength{\fboxsep}{2pt}\colorbox{gray!7}{gray} are commercial models.}
    \label{tab:math_main}
\end{table}

Table~\ref{tab:math_main} presents our main results, comparing \method{} with various baseline methods and commercial systems across the MATH-500, AMC23, and AIME24 benchmarks. We report both Pass@1 and Maj@8 accuracy, as described in \S\ref{sec:results}. 
Overall, \method{} consistently outperforms the baselines that rely solely on outcome-level correctness, such as KTO, IRPO, SimPO, and IPO, as well as simpler methods like RFT. 

For instance, on MATH-500 with the 8B Llama-3.1-Instruct model, \method{} achieves a Pass@1 of 63.2\%, improving from the baseline KTO model’s 60.6\% and substantially surpassing IRPO and RFT. On AMC23, \method{} attains a Pass@1 of 47.5\%, outperforming baselines by a notable margin. On AIME24, where problems require especially intricate multi-step reasoning, \method{} sustains its advantage, demonstrating that the stepwise supervision is particularly valuable for more challenging tasks.
Scaling to the 70B further improves results. \texttt{Llama-3.1-70B-Instruct} with \method{} reaches a Pass@1 of 76.2\% on MATH-500 and continues to excel on AMC23 (60.0\%) and AIME24 (16.7\%). \texttt{Llama-3.3-70B-Instruct} with \method{} model pushes performance higher still, with \method{} achieving 79.6\% on MATH-500, 70.5\% on AMC23, and 29.6\% on AIME24.
Although larger models also benefit from outcome-only alignment techniques, \method{} still delivers consistent gains, indicating that even powerful models trained on extensive data can be further improved by targeting intermediate reasoning quality.
Compared to strong proprietary models, \method{}-enhanced Llama models remain competitive and close the performance gap. For example, while GPT-4o achieves a respectable 73.0\% Pass@1 on MATH-500, O1 series pushes this accuracy to 90.0\% and higher but requires a substantially larger inference budget. In contrast, our \method{}-enhanced \texttt{Llama-3.1-70B-Instruct} model attains 76.2\% Pass@1 on MATH-500 using only a 5k-token budget.

\subsection{Iterative Training}
\begin{table}[t]
    \small
    \centering
    \setlength{\tabcolsep}{7.0pt}
    \begin{NiceTabular}{lcccccc}
         \toprule
         {\bf Method} & \multicolumn{2}{c}{\bf MATH-500} & \multicolumn{2}{c}{\bf AMC23} & \multicolumn{2}{c}{\bf AIME24} \\
         & Pass@1 & Maj@8 & Pass@1 & Maj@8 & Pass@1 & Maj@8 \\
         \midrule
         \rowcolor{blue!7}
         \rowcolor{blue!7} {\small \hspace{-2mm} \emph{Llama-3.1-8B-Instruct} } &  & & &  &  & \\
         \rowcolor{blue!7} Seed model $M_0$ & 53.4 & 55.0 & 35.0 & 37.5 & ~~3.3 & ~~6.7 \\ %
         \midrule
         \rowcolor{blue!7} IPO $M_1$ & 52.6 & 55.8 & 22.5 & 30.0 & ~~3.3 & ~~3.3 \\ %
         \rowcolor{blue!7} SimPO $M_1$ & 55.8 & 57.2 & 25.0 & 25.0 & ~~6.7 & 10.0 \\ %
         \rowcolor{blue!7} Step-DPO $M_1$ & 56.8 & 58.4 & 27.5 & 30.0 & ~~6.7 & 10.0 \\
         \midrule
         \rowcolor{blue!7} Rejection Finetuning $M_1$ & 55.0 & 57.0 & 30.0 & 35.0 & 10.0 & 10.0 \\ %
         \rowcolor{blue!7} Rejection Finetuning $M_2$ & 54.0 & 56.2 & 22.5 & 20.0 & ~~3.3 & ~~6.7 \\ %
         \rowcolor{blue!7} Rejection Finetuning $M_3$ & 53.8 & 56.0 & 30.0 & 32.5 & 10.0 & ~~6.7 \\ %
         \rowcolor{blue!7} IRPO $M_1$ & 58.2 & 59.6 & 35.0 & 35.0 & 10.0 & 10.0 \\ %
         \rowcolor{blue!7} IRPO $M_2$ & 57.2 & 62.4 & 32.5 & 40.0 & ~~6.7 & 10.0 \\ %
         \rowcolor{blue!7} IRPO $M_3$ & 55.4 & 59.2 & 35.0 & 40.0 & ~~6.7 & ~~6.7 \\ %
         \rowcolor{blue!7} KTO $M_1$ & 56.2 & 55.6 & 32.5 & 32.5 & ~~6.7 & 10.0 \\ %
         \rowcolor{blue!7} KTO $M_2$ & 59.4 & 62.8 & 35.5 & 35.0 & \textbf{16.7} & \textbf{16.7} \\ %
         \rowcolor{blue!7} KTO $M_3$ & 60.6 & 61.6 & 35.0 & 32.5 & \textbf{16.7} & \textbf{16.7} \\ %
         \rowcolor{blue!7} \method{} (ours) $M_1$ & 59.4 & 60.6 & 22.5 & 32.5 & 13.3 & 10.0 \\ 
         \rowcolor{blue!7} \method{} (ours) $M_2$ & \textbf{63.6} & 63.0 & 40.0 & 40.0 & 13.3 & \textbf{16.7} \\ %
         \rowcolor{blue!7} \method{} (ours) $M_3$ & 63.2 & \textbf{64.6} & \textbf{47.5} & \textbf{47.5} & \textbf{16.7} & \textbf{16.7} \\ %
         \bottomrule
    \end{NiceTabular}
    \caption{{\bf Iterative training performance} comparing different methods on \texttt{Llama-3.1-8B-Instruct} model. Results show accuracy across multiple iterations ($M_1$, $M_2$, $M_3$) of training on MATH-500, AMC23, and AIME24 test sets using both greedy decoding (Pass@1) and majority voting over 8 samples (Maj@8).}
    \label{tab:iter_ablation}
\end{table}

Table~\ref{tab:iter_ablation} illustrates how model performance evolves over multiple iterative training rounds ($M_1, M_2, M_3$) when starting from the same seed model $M_0$ (\texttt{Llama-3.1-8B-Instruct}). We compare \method{} against other iterative methods such as IRPO, KTO, and Rejection Finetuning.

Overall, \method{} not only achieves higher final performance but also improves more consistently across iterations. For instance, on MATH-500, \method{} progresses from 59.4\% Pass@1 at $M_1$ to 63.2\% at $M_3$, surpassing the gains observed by IRPO and KTO at the same checkpoints. Similarly, on AMC23 and AIME24, \method{} shows steady iterative improvements, reflecting the cumulative value of integrating both process- and outcome-level feedback.
In contrast, Rejection Finetuning (RFT) and IRPO exhibit less stable gains across iterations, with performance sometimes plateauing or even regressing at later rounds. KTO does improve over iterations, but not as robustly as \method{}, highlighting that stepwise feedback adds tangible benefits beyond what outcome-level optimization alone can achieve.

These results underscore the importance of iterative refinement. While simply applying preference-based or rejection-based finetuning may yield some initial improvements, \method{}’s combined stepwise and outcome-level guidance drives steady, sustained enhancements in mathematical reasoning quality, iteration after iteration.

\subsection{Comparison with Step-DPO}
While Step-DPO~\citep{step_dpo} also aims to improve reasoning by focusing on intermediate steps, our method \method{} differs significantly in its approach. Step-DPO identifies erroneous steps and uses rejection sampling to generate correct continuations, requiring substantial computational resources. In contrast, \method{} combines stepwise and outcome-level signals to ensure global solution coherence while remaining computationally efficient. Empirically, Step-DPO shows limited gains after the first iteration ($M_1$), achieving 56.8\% Pass@1 on MATH-500, while \method{} reaches 59.4\%. For implementation, we follow Step-DPO's methodology using \texttt{Llama-3.3-70B-Instruct} for error identification and rejection sampling, and filtering out questions unsolved within 8 attempts. These results demonstrate the advantages of our integrated optimization approach for sustained performance improvements.

\subsection{Preference Optimization Variants}
Table~\ref{tab:iter_ablation} compares \method{} against several baselines after iterative training starting from the 8B seed model $M_0$. Focusing on MATH-500 at $M_1$, \method{} achieves 59.4\% Pass@1—outperforming IPO (52.6\%), SimPO (55.8\%), and even stronger baselines like IRPO (58.2\%) and KTO (56.2\%). On AMC23 and AIME24 at $M_1$, while \method{}’s initial improvements are more modest than IRPO, it remains competitive with other variants and demonstrates stronger subsequent gains. For instance, by $M_3$, \method{} reaches 47.5\% Pass@1 on AMC23, surpassing all baseline methods, and also ties for the highest Pass@1 (16.7\%) on AIME24. 
Collectively, these results underscore the importance of integrating stepwise correctness signals with outcome-level preferences.

\subsection{Evaluating Reasoning Quality}
\begin{table}[h]
    \small
    \centering
    \begin{tabular}{lcc}
        \toprule
        \textbf{8B Model} & \multicolumn{2}{c}{\textbf{Stepwise Errors in Correct Solutions}} \\
        \cmidrule(lr){2-3}
        & \textbf{KTO} & \textbf{\method{}} \\
        \midrule
        $M_0$ & 27.3\% & 27.3\% \\
        $M_1$ & 24.6\% & \textbf{22.9\%} \\
        $M_2$ & 22.6\% & \textbf{20.8\%} \\
        $M_3$ & 21.1\% & \textbf{19.9\%} \\
        \bottomrule
    \end{tabular}
    \caption{{\bf Reasoning Quality Analysis} comparing the ratio of solutions that arrive at correct final answers despite containing erroneous intermediate steps on the MATH-500 test set. Results show that both KTO and \method{} reduce the prevalence of flawed reasoning chains across iterations, with \method{} achieving better consistency.}
    \label{tab:reasoning_quality}
\end{table}

To assess the internal consistency of solutions with correct final answers, 
we evaluate the proportion of solutions that, 
despite having correct final answer $\Rex(\by,\by_{\bx}^{\star})=1$, contain at least one erroneous intermediate step.
We use the ProcessBench \citep{zheng2024processbench} as our evaluation framework, 
which is prompted to identify the earliest error in the generated solution $\by$, 
as detailed in its benchmark construction. 
Additionally, we utilize the critique capabilities of the \texttt{QwQ-32B-Preview} model \citep{qwen2024qwq32b} 
to identify the first error in the reasoning. 
We prompt QwQ using the prompt detailed in Appendix~\ref{app:process_critique_prompt}.
We then measure the percentage of correctly answered problems where QwQ identifies at least one erroneous intermediate step. 

Table~\ref{tab:reasoning_quality} shows the percentage of correctly answered solutions containing errors in reasoning steps, starting from the initial 8B seed model $M_0$, which produces reasoning steps containing errors in 27.3\% 
of its correctly answered solutions on the MATH-500 test set. 
Both \method{} and KTO reduce the prevalence of such errors across iterations, with \method{} showing a greater and more consistent reduction from 27.3\% at $M_0$ to 19.9\% at $M_3$, compared to KTO's more modest improvement to 21.1\%.

\section{Related Work}

\textbf{Outcome-Oriented Methods}
A significant body of work aims to refine LLMs purely based on their final outputs. Large-scale instruction tuning has shown that aligning models with human values or preferences enhances instruction-following capabilities~\citep{ouyang2022training,touvron2023llama}. Outcome-level feedback is often implemented through Reinforcement Learning from Human Feedback (RLHF), as demonstrated by InstructGPT~\citep{ouyang2022training}, or through direct preference optimization techniques such as DPO~\citep{dpo}, KTO~\citep{kto}, SimPO~\citep{simpo} and IPO~\citep{ipo}. These methods optimize the probability of correct or preferred final answers by comparing model-generated candidates against human or synthetic labels. Approaches like RLAIF~\citep{lee2023rlaif} and Constitutional AI~\citep{bai2022constitutional} go further by introducing AI-generated feedback or predefined ethical rules to reduce reliance on human annotations. More recent refinements, such as CGPO~\citep{cgpo}, attempt to improve granularity by providing richer reward signals, though they still primarily judge entire outputs. While effective at improving final-answer correctness, these outcome-focused techniques do not guarantee logical soundness in the intermediate reasoning steps~\citep{thinking_llms}. Models can arrive at correct answers for the wrong reasons, making their solution paths untrustworthy or unfaithful~\citep{unfaithful_explanation_cot,measuring_faithfulness}.

\textbf{Process-Level Feedback and Verification}
Process Reward Models (PRMs)\citep{verify_step_by_step_prm_first_work,orm_prm_training,xiong2024building,omega_prm} focus on stepwise correctness signals. They assign local binary labels or values to each reasoning step, thereby guiding the model to follow logically consistent and provably correct solution trajectories. This paradigm aligns closely with efforts in math reasoning tasks, where datasets like PRM800K\citep{verify_step_by_step_prm_first_work}, CriticBench~\citep{CriticBench}, and ProcessBench~\citep{zheng2024processbench} include detailed reasoning chains and facilitate step-level evaluations. Techniques leveraging PRMs have been integrated into decoding strategies~\citep{li2023inferencetime,chuang2024dola,math_shepherd}, re-ranking~\citep{orm_first}, filtering~\citep{llama3pt1,DBLP:journals/corr/abs-2402-03300}, or iterative improvement loops such as STaR~\citep{star} and ReST~\citep{gulcehre2023reinforced,singh2023beyond}. 
More recent work used synthetic feedback or automatic checks to scale up these stepwise annotations~\citep{math_shepherd,verify_step_by_step_prm_first_work,chiang2024merging,huang-chen-2024-factalign}, showing modest but consistent gains. 
While process-level guidance can refine stepwise correctness, it does not guarantee full alignment with ground-truth solutions. Models may still produce incorrect final outcomes if the reasoning chain fails to converge or if the reward signal is exploited by artificially repeating trivial steps~\citep{DBLP:journals/corr/abs-2410-15115}.

\textbf{Integrating Outcome- and Process-Level Signals}
Recognizing the limitations of relying on only outcome-level or only process-level supervision, recent studies propose combining both signals to align the entire reasoning trajectory with correct and faithful solutions. For example, FactTune~\citep{tian2024finetuning}, and FactAlgin~\citep{huang-chen-2024-factalign} incorporated factuality evaluators and PRMs to produce preference pairs for alignment, demonstrating that mixing final-answer correctness with stepwise verification yields better factual performance.
Similarly, \citet{orm_prm_training} and \citet{DBLP:journals/corr/abs-2402-03300} leveraged feedback for both steps and final outputs to improve math reasoning. 
Although these methods often target general instruction following or long-form factual generation, their principle—using multiple granularities of supervision—holds equally for complex domains like math reasoning. 
Still, many of these approaches face challenges in scaling to very difficult problem sets, balancing the complexities of outcome-level correctness with the subtleties of stepwise coherence, and ensuring that iterative improvements do not plateau prematurely~\citep{bai2022training,DBLP:journals/corr/abs-2312-16682,singh2023beyond}.

\section{Conclusion}

This work proposes \method{}, a training framework that leverages both outcome-level and process-level binary feedback to guide large language models toward more coherent, interpretable, and dependable reasoning. By integrating stepwise correctness signals into the alignment process, \method{} improves the quality of intermediate reasoning steps while maintaining or even enhancing final answer accuracy. Our experiments on challenging mathematical reasoning benchmarks demonstrate consistent gains in performance, particularly under iterative training and for complex reasoning tasks. These findings underscore the value of aligning not only final outcomes but also the entire reasoning trajectory. We envision \method{} as a stepping stone toward more reliable reasoning in LLMs.

\section*{Limitations}

While our \method{} framework shows promise in improving both outcome-level correctness and the internal coherence of reasoning steps, several limitations remain.

First, outcome-level feedback signals can be noisy or imperfect. In mathematical reasoning, even automatically verified final answers may occasionally fail to capture all nuances of correctness. For instance, subtle formatting differences or unconventional but valid representations might lead to false negatives. This noise can limit the precision of the training signal, potentially hindering further improvements.

Second, our approach currently relies on access to ground-truth solutions for both final answers and (implicitly) for guiding stepwise correctness. In scenarios where no high-quality ground-truth reasoning paths are available, or where the notion of \textit{correct} intermediate reasoning steps is inherently ambiguous, it may be challenging to define meaningful stepwise feedback. Developing methods that can learn from weaker or noisier references, or even from purely preference-based evaluations without explicit ground-truth reasoning, remains an open problem.

Finally, our experiments focus on cases where at least some correct outcomes or partially correct steps are achievable. If the outcome is always incorrect and the model struggles to produce even partially valid intermediate steps, it is unclear whether \method{} would effectively bootstrap performance. In such difficult settings, additional techniques—such as curriculum learning, stronger initialization, or tailored exploration strategies—may be necessary before stepwise feedback can meaningfully guide the model toward correct final answers and improved reasoning processes.

\bibliographystyle{assets/plainnat}
\bibliography{paper,orm_prm,fa_bib}

\clearpage
\newpage
\beginappendix

\section{Decontamination}
To prevent data leakage between training and test sets, we perform standard decontamination by normalizing text (converting to lowercase and removing non-alphanumeric characters) and checking for exact string matches between test questions and training prompts \citep{llama3pt1}. We remove any matching examples from the training data. This process is applied to all datasets in our evaluation.
Even if mild contamination were present, we expect any resulting performance inflation to be small and consistent across all conditions, leaving the relative comparisons between our methods largely unaffected.

\section{Details of API Usage for Proprietary Models}
\label{app:api_usage_details}
In our experiments, we evaluated several proprietary models via their respective APIs: O1 (metrics are self-reported), O1-Mini (\texttt{o1-mini-2024-09-12}, MATH-500 is self-reported \footnote{Numbers from \url{https://github.com/openai/simple-evals}}), Gemini 1.5 Pro (\texttt{gemini-1.5-pro-002}), GPT-4o (\texttt{gpt-4o-2024-08-06}), Claude 3.5 Sonnet (\texttt{claude-3-5-sonnet-20241022}), and Grok-Beta. 
These experiments took place on November 15 and 16, 2024. 
For each model, questions were used directly as user prompts. 
For greedy decoding, we set the temperature to 0.0 to ensure deterministic outputs, except for o1 models where we used temperature 1.0 due to API restrictions (only temperature 1.0 is allowed) and took the first sample.
For sampling, we set the temperature to 0.7 and performed 8 generations per question to enable majority voting.
\begin{tcolorbox}[colback=orange!5!white, colframe=orange!80!black, title=Response Generation for Proprietary Models]
\textbf{User:}
\begin{verbatim}
Please answer the following question step-by-step. Once you have the final answer, place it
on a new line as: The final answer is \$\boxed{answer}\$.
Question: {{ question }}
\end{verbatim}
\end{tcolorbox}

\section{Prompts}
\label{app:prompts}
Prompt templates \footnote{The prompt template was from \url{https://huggingface.co/datasets/meta-llama/Llama-3.1-70B-Instruct-evals}} for generating solutions are given below in \cref{app:solution_template}.

\begin{tcolorbox}[colback=blue!5!white, colframe=blue!80!black, title=Response Generation Template from \cite{llama3pt1}]
\label{app:solution_template}
\textbf{User:}
\begin{verbatim}
Solve the following math problem efficiently and clearly:

For simple problems (2 steps or fewer):
Provide a concise solution with minimal explanation.

For complex problems (3 steps or more):
Use this step-by-step format:

## Step 1: [Concise description]
[Brief explanation and calculations]

## Step 2: [Concise description]
[Brief explanation and calculations]

...

Regardless of the approach, always conclude with:

Therefore, the final answer is: \$\boxed{answer}\$. I hope it is correct.

Where [answer] is just the final number or expression that solves the problem.

Problem: {{ question }}
\end{verbatim}
\end{tcolorbox}

Prompt for \texttt{Llama-3.1-70B-Instruct} to provide stepwise feedback on candidate solutions $\by$. The model analyzes each step $s_h$ of a potential solution against the correct answer $\by^\star$, evaluating the reasoning and accuracy of each step. The feedback is structured in JSON format with fields for step number, reflection on the reasoning, and a binary decision on whether the step contributes positively to reaching the solution.

\begin{tcolorbox}[colback=green!5!white, colframe=green!80!black, title=Generation Prompt for Stepwise Feedback]
\label{app:stepwise_feedback_prompt}
\textbf{User:} 
\footnotesize
\begin{verbatim}
Please analyze the following problem and its potential solution step-by-step.
Provide feedback on each step and determine if it contributes positively to reaching the 
correct solution.

<problem>
{{ problem }}
</problem>

<correct solution>
{{ answer }}
</correct solution>

<potential answer>
{%
<step {{ loop.index }}>
{{ step }}
</step {{ loop.index }}>
{%
</potential answer>

Analyze your **potential solution** as if you had originally generated it. 
Carefully review each step, considering its reasoning, accuracy, and execution. 
Assess whether the step contributes positively to reaching the correct solution. 
Where necessary, refine the step to address any flaws or gaps. Use the correct answer 
as a ground truth reference to guide your analysis.

Provide your output in JSON format, where each element represents a step of the solution.
Use the fields below:

- **step**: The step order number in the reasoning process.
- **reflection**: A concise evaluation of the accuracy of the reasoning in this step 
(point out why it helps or hinders the solution).
- **decision**: The evaluation of the step, either "positive" or "negative".

The expected output format follows:

```json
[
    {
        "step": 1,
        "reflection": "[evaluation of step 1 reasoning]",
        "decision": "positive"
    },
    {
        "step": 2,
        "reflection": "[evaluation of step 2 reasoning]",
        "decision": "negative"
    },
    ...\\   
]
```

# Notes

- Assign **"negative"** only to steps that are clearly incorrect and prevent the 
solution from progressing.
- Use the correct answer as one of the ground truth references.
- Ensure the reflections remain concise, with the total explanation under 200 words 
across all steps combined.
\end{verbatim}
\end{tcolorbox}

\begin{tcolorbox}[colback=yellow!5!white, colframe=yellow!80!black, 
    title=Process Critique Prompt from ProcessBench \citep{zheng2024processbench}]
\textbf{User:}
\begin{verbatim}
The following is a math problem and a solution (split into paragraphs, enclosed with tags and 
indexed from 0):

<problem>
{{ problem }}
</problem>

<solution>
{%
<step {{ loop.index }}>
{{ step }}
</step {{ loop.index }}>
{%
</solution>

Your task is to review and critique the solution paragraph by paragraph. Once you identify an 
error in a paragraph, return the index of the paragraph where the earliest error occurs. 
Otherwise, return the index of -1 (which typically denotes "not found").

Please put your final answer (i.e., the index) in \boxed{{}}.
\end{verbatim}
\label{app:process_critique_prompt}
\end{tcolorbox}

\section{Qualitative Examples}

We analyze several examples from \texttt{Llama-3.3-70B-Instruct} \method{} $M_1$ on MATH-500 to understand how Step-KTO helps improve mathematical reasoning. The examples demonstrate three key scenarios where Step-KTO provides effective feedback: (1) when all steps and the final answer are correct, (2) when intermediate steps contain errors but lead to the correct final answer, and (3) when both intermediate steps and the final answer are incorrect.

\begin{tcolorbox}[colback=color2!5!white, colframe=color2!80!black, 
    title=Step-KTO Generation Example 1: Final Answer is Right and All Steps are Right]
\footnotesize
\begin{verbatim}
problem: For a point $P,$ let $d_1,$ $d_2$ and $d_3$ represents the distances from $P$ 
         to the planes $x - z = 0,$ $x - 2y + z = 0,$ and $x + y + z = 0.$  Let $S$ be 
         the set of points $P$ such that\n\\[d_1^2 + d_2^2 + d_3^2 = 36.\\]Find the region 
         of the volume enclosed by $S.$
answer: "288 \\pi",
subject: "Precalculus",
level: 4,
unique_id: "test/precalculus/1056.json"

Model prediction:
## Step 1: Understand the problem and the equation given
The problem provides an equation $d_1^2 + d_2^2 + d_3^2 = 36$ where $d_1,$ $d_2,$ and 
$d_3$ represent the distances from a point $P$ to the planes $x - z = 0,$ $x - 2y + z 
= 0,$ and $x + y + z = 0$ respectively. We need to find the region of the volume enclosed by 
the set $S$ of points $P$ that satisfy this equation.

## Step 2: Calculate the distance of a point to each plane
The distance $d$ from a point $(x_0, y_0, z_0)$ to a plane $Ax + By + Cz + D = 0$ is given by 
the formula $d = \frac{|Ax_0 + By_0 + Cz_0 + D|}{\sqrt{A^2 + B^2 + C^2}}$. Applying this formula 
to each plane, we get:
- $d_1 = \frac{|x - z|}{\sqrt{1^2 + (-1)^2}} = \frac{|x - z|}{\sqrt{2}}$
- $d_2 = \frac{|x - 2y + z|}{\sqrt{1^2 + (-2)^2 + 1^2}} = \frac{|x - 2y + z|}{\sqrt{6}}$
- $d_3 = \frac{|x + y + z|}{\sqrt{1^2 + 1^2 + 1^2}} = \frac{|x + y + z|}{\sqrt{3}}$

## Step 3: Substitute the distance formulas into the given equation
Substituting $d_1,$ $d_2,$ and $d_3$ into the equation $d_1^2 + d_2^2 + d_3^2 = 36$, we get:
$\left(\frac{x - z}{\sqrt{2}}\right)^2 + \left(\frac{x - 2y + z}{\sqrt{6}}\right)^2 + 
\left(\frac{x + y + z}{\sqrt{3}}\right)^2 = 36$

## Step 4: Simplify the equation
Expanding and simplifying, we have:
$\frac{(x - z)^2}{2} + \frac{(x - 2y + z)^2}{6} + \frac{(x + y + z)^2}{3} = 36$

## Step 5: Further simplification and rearrangement
Multiplying through by 6 to clear the denominators gives:
$3(x - z)^2 + (x - 2y + z)^2 + 2(x + y + z)^2 = 216$

## Step 6: Expand the terms
Expanding each term yields:
$3(x^2 - 2xz + z^2) + (x^2 - 4xy + 4y^2 + z^2 + 2xz - 4yz) + 2(x^2 + 2xy + 2xz + y^2 + 2yz + z^2) = 216$

## Step 7: Combine like terms
Combining like terms, we get:
$3x^2 - 6xz + 3z^2 + x^2 - 4xy + 4y^2 + z^2 + 2xz - 4yz + 2x^2 + 4xy + 4xz + 2y^2 + 4yz + 2z^2 = 216$

## Step 8: Simplify further
This simplifies to:
$6x^2 + 6y^2 + 6z^2 + 0xy + 0xz + 0yz = 216$

## Step 9: Divide by 6
Dividing the entire equation by 6 gives:
$x^2 + y^2 + z^2 = 36$

## Step 10: Interpret the result
The equation $x^2 + y^2 + z^2 = 36$ represents a sphere with radius 6 centered at the origin (0,0,0).

The final answer is: $\boxed{288\pi}$
\end{verbatim}
\end{tcolorbox}

\begin{tcolorbox}[colback=color1!5!white, colframe=color1!80!black, 
    title=Step-KTO Generation Example 2: Final Answer is Right but Step 2 is Wrong]
\small
\begin{verbatim}
problem: For how many real values of $\sqrt{120-\sqrt{x}}$ is an integer?
answer: "11"
subject: "Algebra"
level: 5
unique_id: "test/algebra/1282.json"

Model prediction:
## Step 1: Understand the problem and identify the condition for $\sqrt{120-\sqrt{x}}$ to be an integer.
We need $\sqrt{120-\sqrt{x}}$ to be an integer, which means $120-\sqrt{x}$ must be a perfect square.

## Step 2: List all perfect squares less than or equal to 120.
The perfect squares less than or equal to 120 are: 0, 1, 4, 9, 16, 25, 36, 49, 64, 
81, 100, and 121. However, since we are considering $120-\sqrt{x}$, the relevant 
perfect squares are from 0 to 120, which are 0, 1, 4, 9, 16, 25, 36, 49, 64, 81, 100, 
and 121 is not included because it exceeds 120.

## Step 3: Calculate the possible values of $\sqrt{x}$.
For each perfect square $y$ from the list, we calculate $\sqrt{x} = 120 - y$. This 
gives us the possible values of $\sqrt{x}$ as: 120, 119, 116, 111, 104, 95, 84, 71, 56, 39, 20.

## Step 4: Consider the validity of each possible value of $\sqrt{x}$.
Since $\sqrt{x}$ cannot be negative and must be real, all calculated values in step 
3 are valid because they are all positive or zero. However, we should also consider 
that $\sqrt{x}$ itself must be real, meaning $x \geq 0$. Given that $\sqrt{x} = 120 
- y$, the smallest value of $\sqrt{x}$ is when $y = 100$, giving $\sqrt{x} = 20$. 
For $y = 121$, $\sqrt{x}$ would be negative, which is not valid for real $x$. Thus, 
we only consider perfect squares up to 100.

## Step 5: Count the number of valid values of $\sqrt{x}$ that make 
$\sqrt{120-\sqrt{x}}$ an integer.
From step 3 and considering the correction in step 4, we see that for each perfect 
square up to 100 (inclusive), we get a valid $\sqrt{x}$. Thus, we count the number 
of perfect squares from 0 to 100, which are 11 (0, 1, 4, 9, 16, 25, 36, 49, 64, 81, 
100).

The final answer is: $\boxed{11}$

QwQ Critique:
\boxed{2}

The error in the solution occurs in the 2nd paragraph, where the reasoning incorrectly 
states "The perfect squares less than or equal to 120 are: 0, 1, 4, 9, 16, 25, 36, 49, 
64, 81, 100, and 121." It's incorrect because 121 is greater than 120, not less than or 
equal to, and shouldn't be included in the list.
\end{verbatim}
\end{tcolorbox}

\begin{tcolorbox}[colback=red!5!white, colframe=red!80!black, 
    title=Step-KTO Generation Example 3: Final Answer is Wrong]
\footnotesize
\begin{verbatim}
Problem: Find the greatest integer less than $(\sqrt{7} + \sqrt{5})^6.$  
         (Do not use a calculator!)
answer: "13535"
subject: "Intermediate Algebra"
level: 3
unique_id: "test/intermediate_algebra/607.json"

Model prediction:
## Step 1: Understand the problem and identify the key elements
We need to find the greatest integer less than $(\sqrt{7} + \sqrt{5})^6$ without using a
calculator. This involves expanding the given expression and simplifying it to a form where 
we can estimate its value.

## Step 2: Apply the Binomial Theorem to expand the expression
The Binomial Theorem states that for any non-negative integer $n$, \((a + b)^n = 
\sum_{k=0}^{n} \binom{n}{k}a^{n-k}b^k\). Applying this to our problem with $a = 
\sqrt{7}$, $b = \sqrt{5}$, and $n = 6$, we get \((\sqrt{7} + \sqrt{5})^6 = 
\binom{6}{0}(\sqrt{7})^6 + \binom{6}{1}(\sqrt{7})^5(\sqrt{5}) + 
\binom{6}{2}(\sqrt{7})^4(\sqrt{5})^2 + \binom{6}{3}(\sqrt{7})^3(\sqrt{5})^3 + 
\binom{6}{4}(\sqrt{7})^2(\sqrt{5})^4 + \binom{6}{5}(\sqrt{7})(\sqrt{5})^5 + 
\binom{6}{6}(\sqrt{5})^6\).

## Step 3: Calculate each term of the expansion
Calculating each term:
- \(\binom{6}{0}(\sqrt{7})^6 = 1 \cdot 7^3 = 343\)
- \(\binom{6}{1}(\sqrt{7})^5(\sqrt{5}) = 6 \cdot 7^2 \cdot 7 \cdot \sqrt{5} = 6 
\cdot 49 \cdot 7 \cdot \sqrt{5} = 2058\sqrt{5}\)
- \(\binom{6}{2}(\sqrt{7})^4(\sqrt{5})^2 = 15 \cdot 7^2 \cdot 5 = 15 \cdot 49 \cdot 5 
= 3675\)
- \(\binom{6}{3}(\sqrt{7})^3(\sqrt{5})^3 = 20 \cdot 7 \cdot 7 \cdot 5 \cdot \sqrt{5} 
= 20 \cdot 49 \cdot 5 \cdot \sqrt{5} = 4900\sqrt{5}\)
- \(\binom{6}{4}(\sqrt{7})^2(\sqrt{5})^4 = 15 \cdot 7 \cdot 5^2 = 15 \cdot 7 \cdot 25 
= 2625\)
- \(\binom{6}{5}(\sqrt{7})(\sqrt{5})^5 = 6 \cdot \sqrt{7} \cdot 5^2 \cdot \sqrt{5} = 
6 \cdot \sqrt{7} \cdot 25 \cdot \sqrt{5} = 150\sqrt{35}\)
- \(\binom{6}{6}(\sqrt{5})^6 = 1 \cdot 5^3 = 125\)

## Step 4: Combine like terms and simplify
Combining like terms, we notice that terms with \(\sqrt{5}\) and \(\sqrt{35}\) will 
contribute to non-integer values, but we can simplify the expression by recognizing 
that the integer part of the expansion will come from the terms without square roots 
(since they are multiplied by integers) and the terms with square roots will contribute 
to the fractional part. The terms without square roots are \(343 + 3675 + 2625 + 125 = 
6768\).

## Step 5: Estimate the value of the expression with square roots
To estimate the value of the terms with square roots, we note that \(2058\sqrt{5} 
\approx 2058 \cdot 2.236 = 4599.33\), \(4900\sqrt{5} \approx 4900 \cdot 2.236 = 
10943.4\), and \(150\sqrt{35} \approx 150 \cdot 5.916 = 886.4\). Summing these 
approximations gives us an estimate of the non-integer contributing terms: \(4599.33 + 
10943.4 + 886.4 \approx 15429.13\).

## Step 6: Combine the integer and non-integer parts
Adding the integer part (from step 4) to the estimated value of the non-integer part 
(from step 5), we get \(6768 + 15429.13 \approx 22197.13\).

## Step 7: Determine the greatest integer less than the result
The greatest integer less than \(22197.13\) is \(22197\).

The final answer is: $\boxed{22197}$

\end{verbatim}
\end{tcolorbox}

\end{document}